\documentclass[11pt]{article}

\usepackage[preprint]{acl}

\usepackage{times}
\usepackage{latexsym}

\usepackage[T1]{fontenc}
\usepackage[utf8]{inputenc}

\usepackage{microtype}

\usepackage{inconsolata}

\usepackage{graphicx}
\usepackage{subcaption}

\usepackage{amsmath}
\usepackage{amssymb}
\usepackage{booktabs}
\usepackage{makecell}
\usepackage{multirow}
\title{ClaHF: A Human Feedback-inspired Reinforcement Learning Framework for Improving Classification Tasks}

\author{
 \textbf{Tianxiang Xu\textsuperscript{1}},
 \textbf{Xiaoyan Zhu\textsuperscript{1}},
 \textbf{Xin Lai\textsuperscript{1}},
 \textbf{Jiayin Wang\textsuperscript{1}},
\\
 \textsuperscript{1}School of Computer Science and Technology, Xi'an Jiaotong University
\\
 \small{
   \textbf{Correspondence:} \href{wangjiayin@mail.xjtu.edu.cn}{wangjiayin@mail.xjtu.edu.cn}
 }
}

\begin{document}
\maketitle
\begin{abstract}
Text classification models are typically trained via supervised fine-tuning (SFT).
However, SFT essentially performs behavior cloning from instance-wise labels and thus fails to adequately capture relative preference relations among samples, which limits the model’s ability to shape decision boundaries and calibrate predictive confidence.
In this paper, we propose ClaHF, a human feedback-inspired reinforcement learning (RL) framework for text classification that integrates preference modeling and RL optimization into the classification pipeline without requiring additional human annotations.
Unlike prior work that relies solely on instance-wise supervision, ClaHF constructs multiple candidate predictions together with their relative ranking relations, and jointly models the Top-1 preference and the ordering among non-optimal candidates within a reward model (RM).
This design converts conventional label supervision into preference signals that are directly applicable to policy optimization.
We conduct systematic evaluations on eight classification tasks spanning three categories of scenarios.
Results demonstrate that ClaHF consistently improves both classification performance and confidence calibration across diverse language models (LMs). The data and code are available at \url{https://anonymous.4open.science/r/ClaHF}.
\end{abstract}

\section{Introduction}

Text classification is one of the most fundamental and important tasks in natural language processing (NLP), forming the basis of a wide range of downstream applications such as sentiment analysis, news categorization and emotion recognition \citep{li2022survey}.
In recent years, the rapid development of pre-trained LMs has led to remarkable improvements in these domains, substantially boosting downstream performance \citep{sun2023text}.
Nevertheless, most existing text classification models are still trained via SFT, where model predictions are aligned with gold labels on an instance-by-instance basis \citep{fonseca2025instance,wei2025protolens}.
From a learning paradigm perspective, SFT essentially performs behavior cloning: the model is trained to imitate annotators’ decisions on the training data in order to learn classification boundaries \citep{chu2025sft}.

Although SFT effectively activates the semantic representation capacity of pre-trained models, it lacks further exploration of the decision space \citep{fu2023stability}.
When data quality is limited, when class boundaries differ only in subtle semantics, or when annotations themselves contain bias, SFT tends to amplify these issues.
As a result, the model may overfit the label distribution rather than the underlying semantic patterns, leading to misclassification and unreliable predictions \citep{ye2025analyzing}.
This phenomenon shares a similar root cause with the amplification of hallucinations in generative tasks, but in the classification setting it manifests as unstable decision boundaries and degraded generalization \citep{almeida2021mitigating}.

As a complementary paradigm, Reinforcement learning from human feedback (RLHF) \citep{stiennon2020learning,ouyang2022training} has demonstrated great potential in aligning LMs with human preferences, particularly in generative tasks such as dialogue and instruction following \citep{dong2024rlhf,ouyang2025towards,bai2022training}.
Instead of predicting a single target, RLHF learns from preference signals and optimizes outputs through relative comparisons, thereby better reflecting subjective human judgments \citep{jiang2025survey}.
This naturally raises the question: can the principles of RLHF be transferred to classification tasks to overcome the inherent limitations of SFT?

\begin{figure*}[t]
  \centering
  \begin{subfigure}{0.63\linewidth}
    \centering
    \includegraphics[width=\linewidth]{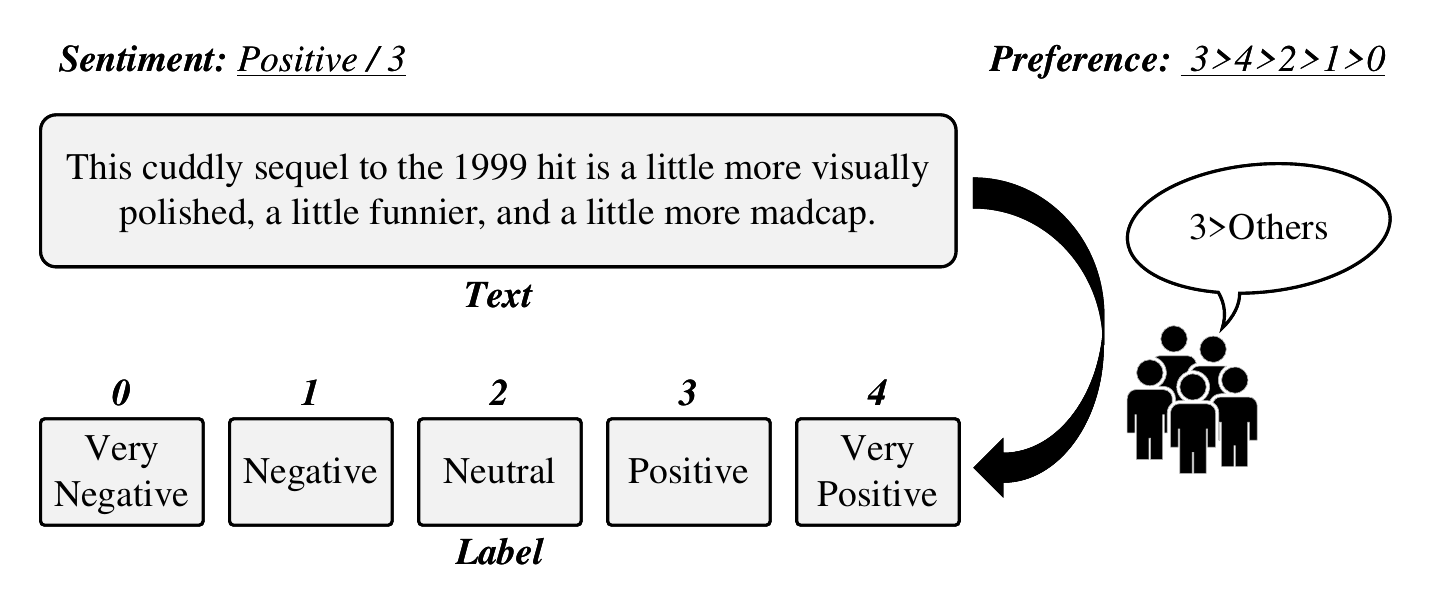}
    \caption{Implicit preference signals in human decisions.}
    \label{fig:1-a}
  \end{subfigure}
  \hfill
  \begin{subfigure}{0.33\linewidth}
    \centering
    \includegraphics[width=\linewidth]{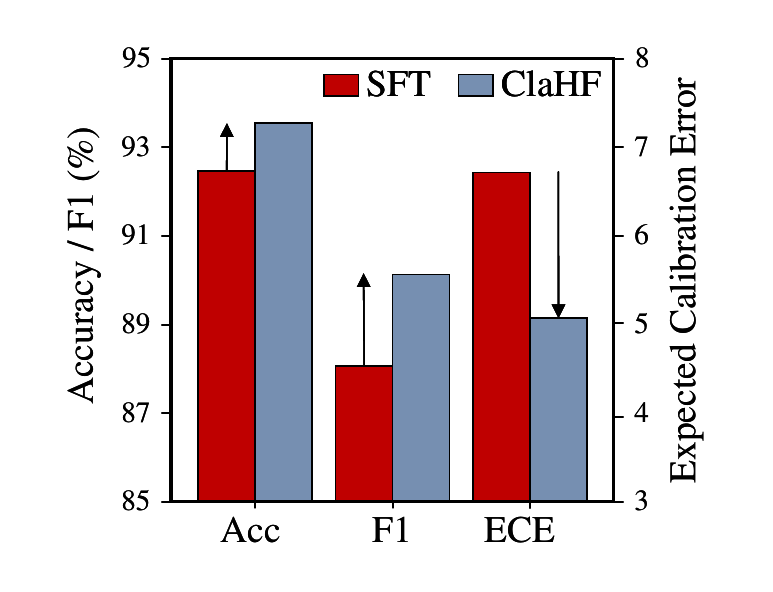}
    \caption{Improved text classification.}
    \label{fig:1-b}
  \end{subfigure}
  \caption{(a) An example illustrating the implicit preference signal in the construction of the SST-5 sentiment classification dataset.
  (b) The performance improvements of ClaHF on text classification tasks (\emph{e.g.}, higher accuracy and F1, lower error).
  More results are reported in Section~\ref{sec:4.2}.}
  \label{fig:1}
\end{figure*}

Unfortunately, directly extending RLHF to classification faces fundamental challenges.
First, unlike the rich sequential outputs of generative models, the output space of classification is discrete and lacks expressive structure \citep{casper2023open}.
Second, conventional RLHF relies heavily on manually annotated preference datasets, whose construction is costly and difficult to reuse across tasks and label spaces \citep{casper2023open,lindstrom2024ai}.
These issues make existing RLHF techniques hard to apply to classification, and relevant studies remain scarce.

To address these challenges, we propose ClaHF, a novel human feedback-inspired RL framework for text classification that introduces RLHF into classification training without requiring any additional human annotation.
ClaHF is motivated by a key insight: although standard classification datasets do not contain explicit preference annotations, the ground-truth label of each instance implicitly encodes the preference signal underlying human decisions (Figure~\ref{fig:1}(\subref{fig:1-a})).
Based on this insight, ClaHF automatically constructs multiple candidate predictions and their ranking relations from raw classification data, yielding preference data at no extra cost.
These preference signals are then used to train a RM that evaluates the relative quality of different predictions.
Finally, the classification model is optimized with a preference-aligned proximal policy optimization (PPO) \citep{schulman2017proximal} algorithm.
Notably, ClaHF is applicable to both binary and multi-class classification.

Unlike conventional instance-wise supervision, ClaHF leverages a multi-stage collaborative process that enables the model not only to ``predict correctly'' but also to ``align with preference signals'' embedded in the data, thereby overcoming the limitations of SFT in classification.
We conduct systematic experiments on eight benchmark datasets covering binary, multi-class, and code classification tasks.
The results show that ClaHF significantly improves both performance and decision reliability over strong SFT baselines (Figure~\ref{fig:1}(\subref{fig:1-b})).
These findings demonstrate that ClaHF bridges the gap between RLHF and classification, and introduce a new perspective: classification models can benefit from preference alignment in addition to label supervision.

\section{Related Work}

\textbf{Supervised Learning for Classification.}
The dominant paradigm is to perform SFT on labeled data using pre-trained models such as RoBERTa \citep{liu2019roberta} and Qwen3 \citep{yang2025qwen3}, which yields strong discriminative capability.
Despite its empirical success, this training paradigm inherits the limitations of instance-wise supervision, which often leads to overconfidence, unreliable decisions, and misclassification of ambiguous samples near class boundaries.
To alleviate these issues, a number of studies have proposed improved objectives.
\citet{lin2017focal} introduced focal loss to emphasize hard and minority samples, while \citet{muller2019does} proposed label smoothing to improve generalization by injecting soft targets and noise regularization.
However, these methods still treat labels as independent supervision signals and fail to exploit the latent relative preferences and ranking relations among samples.
Consequently, the training process remains essentially behavior cloning, which is insufficient to provide deeper guidance at the decision level.

\textbf{Applications of RLHF.}
RLHF was originally developed to optimize agent behavior in simulated environments \citep{christiano2017deep}, and was later applied to domains such as autonomous driving \citep{wu2022prioritized} and robot learning \citep{wang2022skill}.
Prior work \citep{xu2023imagereward,zhang2024hive,fang2025human} has also demonstrated the effectiveness of human feedback in image generation.
More recently, the successful application of RLHF to large language models (LLMs) has substantially advanced NLP \citep{stiennon2020learning,ouyang2022training,ziegler2019fine}.
By modeling human preferences and performing policy updates with PPO, RLHF integrates reward-driven mechanisms into language modeling, fundamentally reshaping dialogue reasoning, safety alignment, and long-form generation \citep{ji2025survey,liu2023summary}.
However, within downstream NLP tasks, RLHF has so far been largely confined to generative scenarios, such as high-quality code test case generation \citep{steenhoek2025reinforcement}.

\textbf{Preference Learning.}
Preference learning is a core component of RLHF and models relations among samples via pairwise or listwise comparisons.
It has been widely adopted in ranking and recommendation systems \citep{furnkranz2010preference}.
In NLP, preference learning is mainly used for generative tasks, although recent studies \citep{atapour2021rank,kim2023prefer} have attempted to introduce preference modeling into classification to overcome the limitations of instance-wise supervision.
For example, \citet{kim2023prefer} proposed the P2C method, which incorporates preference relations during training to enhance the model’s discrimination near class boundaries.
Nevertheless, existing approaches typically treat preference ranking as an auxiliary loss jointly optimized with SFT; they neither construct an independent RM nor perform policy optimization with RL, and thus remain within the supervised learning paradigm.

\section{Method}
In this section, we present ClaHF, a novel human feedback-inspired framework for classification.
Unlike conventional training paradigms that rely solely on instance-wise supervision, ClaHF borrows the core ideas of RLHF and introduces preference signals into text classification through reward modeling and RL, enabling finer calibration of the predictive distribution and improving both classification performance and decision reliability.

\subsection{Problem Definition}

We focus on text classification tasks, including both binary and multi-class settings.
Given a dataset $\mathcal{D} = \{(x_i,y_i)\}_{i=1}^{N}$, where $x_i$ denotes an input sequence (typically a sentence or a paragraph) and $y_i$ is the corresponding class label, conventional approaches learn a classifier $f_\theta$ by minimizing instance-wise losses to maximize the agreement between predictions and ground-truth labels.
However, such instance-wise supervision provides only a single discriminative signal and cannot characterize the relative quality among different predictions, which limits further optimization.

\begin{figure*}[t]
  \centering
  \includegraphics[width=0.99\linewidth]{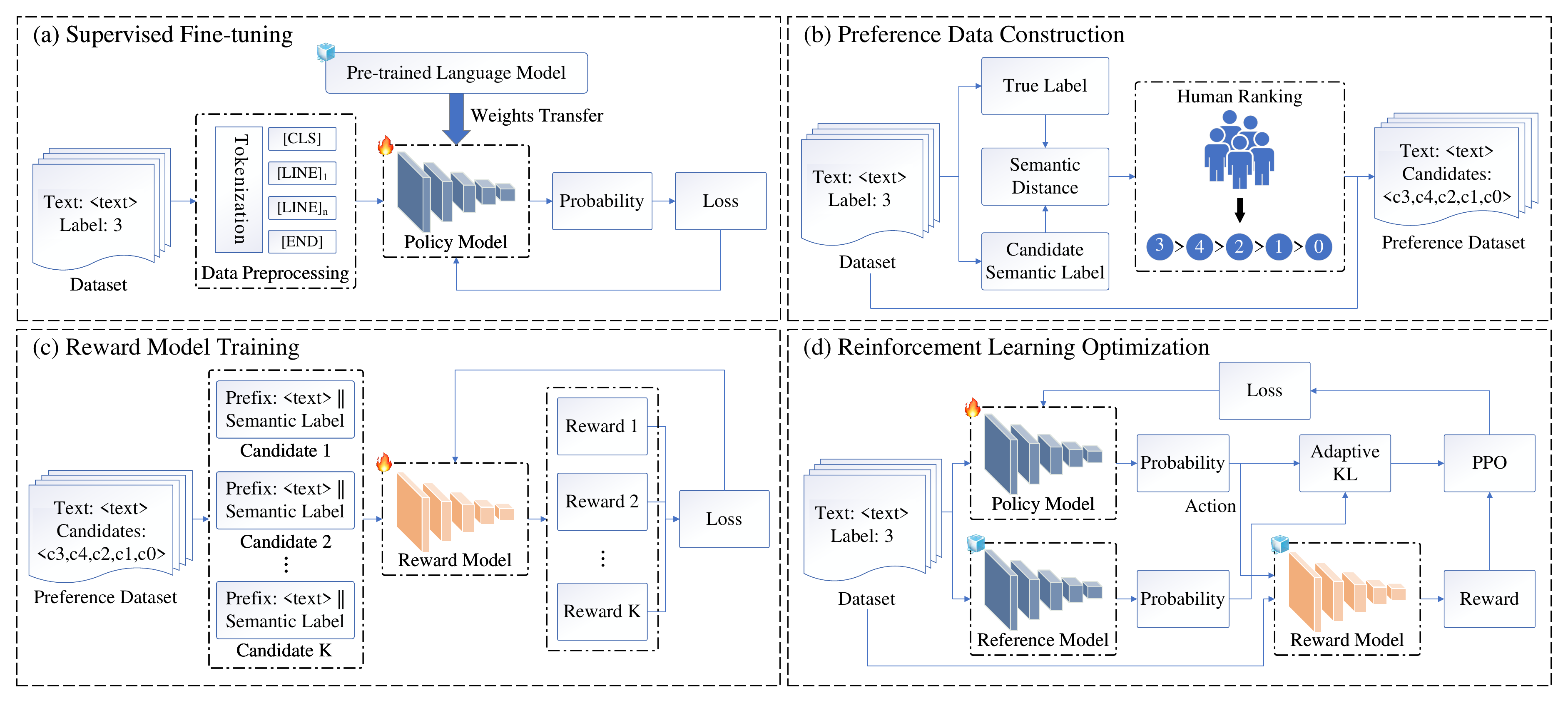}
  \caption{Overall framework of ClaHF.
  (a) SFT to provide high-quality initialization.
  (b) Automatic construction of preference data from the original classification dataset.
  (c) Training the RM with preference data.
  (d) RL optimization of the policy model using the trained RM.}
  \label{fig:2}
\end{figure*}

To this end, we propose ClaHF, which augments standard label supervision with automatically constructed preference signals.
Specifically, ClaHF introduces ranking constraints among multiple candidate predictions for the same instance to capture fine-grained differences in model outputs.
The goal of ClaHF is therefore to learn a classifier $f_\theta$ that not only predicts the correct label but also optimizes its predictive distribution under preference constraints, thereby aligning the model with implicit ``human feedback'' embedded in the data.
The overall framework of ClaHF is illustrated in Figure~\ref{fig:2}.

\subsection{Supervised Fine-tuning}

The SFT stage provides a stable and performant initial policy model $f_\theta$ for subsequent RL.
In ClaHF, the model $f_\theta$ consists of a pre-trained LM $g_\phi$ and a randomly initialized classification head $W_\psi$.
Given an input text $x_i$, we encode it into token sequences and feed them into $f_\theta$ to obtain hidden representations, followed by a softmax layer to produce the class probability distribution:
\begin{equation}
  \label{eq:1}
  p_{\theta}(y_i \, | \, x_i) = \operatorname{Softmax}(f_\theta(x_i)).
\end{equation}

We optimize the model using the standard cross-entropy loss:
\begin{equation}
  \label{eq:2}
  \mathcal{L}_{\text{sft}} = -\mathbb{E}_{(x_i,y_i) \sim \mathcal{D}} \log p_{\theta}(y_i \, | \, x_i).
\end{equation}

This stage ensures that the policy model has strong baseline classification capability.
However, the supervision signal is purely instance-wise and only enforces consistency with the ground-truth label, without reflecting the relative quality among alternative predictions.

\subsection{Preference Data Construction}

Training the RM requires data annotated with preference signals. In conventional RLHF, such data are manually labeled, which is expensive and task-specific. In contrast, classification datasets naturally contain ground-truth labels that implicitly encode human preferences. Based on this observation, ClaHF introduces a fully automated method for constructing preference data without any additional annotation.

Specifically, given a standard $K$-class dataset $\mathcal{D} = \{(x_i, y_i)\}_{i=1}^{N}$, we first define a set of semantic label templates:
\begin{equation}
  \label{eq:3}
  \mathcal{Y}_{\text{text}} = \{0:c_{0}, 1:c_{1}, \dots, K-1:c_{K-1}\},
\end{equation}
where each $c_j$ is a natural language description of class $j$, referred to as a candidate semantic label.
Examples for SST-5 are shown in Table~\ref{tab:1}, and those for other datasets are provided in Appendix ~\ref{sec:apB}.

\begin{table}
\setlength\tabcolsep{3pt}
  \centering
  \small
  \begin{tabular}{c|c|c}
    \toprule
    \textbf{Dataset}    & \textbf{Label} & \textbf{Candidate Semantic Label} \\
    \midrule
    \multirow{5}{*}{SST-5}
                &  0    &  Prediction: the sentence is very negative  \\
                &  1    &  Prediction: the sentence is negative     \\
                &  2    &  Prediction: the sentence is neutral        \\
                &  3    &  Prediction: the sentence is positive       \\
                &  4    &  Prediction: the sentence is very positive   \\
    \bottomrule
  \end{tabular}
  \caption{
    Ground-truth labels and corresponding candidate semantic labels for SST-5.
  }
  \label{tab:1}
\end{table}

For each instance $(x_i,y_i)$, we rank all candidate semantic labels $c_j$ in ascending order of their semantic distance to the true label $y_i$, yielding a candidate list:
\begin{equation}
  \label{eq:4}
  C_i = \{c_{i1}, c_{i2}, \dots, c_{iK}\},
\end{equation}
where $c_{i1}$ always corresponds to the ground-truth class, and the remaining candidates are ordered by increasing semantic distance.
For example, in SST-5, if the true label is 3, the resulting list is $\{c_{3}, c_{4}, c_{2}, c_{1}, c_{0}\}$.

Aggregating all instances produces the preference dataset $\mathcal{D}_{\text{pre}} = \{(x_i, C_i)\}_{i=1}^{N}$.
Unlike the original dataset that only contains binary correctness information, $\mathcal{D}_{\text{pre}}$ encodes relative plausibility among classes and naturally introduces a ``Top-1 is better than others'' preference structure.

\subsection{Reward Model Training}

The third stage of ClaHF trains a RM to evaluate the relative quality of text–candidate pairs and provide reward signals for subsequent policy optimization.
The RM $f_{\text{rm}}$ consists of the same pre-trained encoder $g_\phi$ as in SFT and a linear scoring head that outputs a scalar reward.

Given $\mathcal{D}_{\text{pre}} = \{(x_i, C_i)\}_{i=1}^{N}$, for each $(x_i, C_i)$ we concatenate $x_i$ with each $c_{ij}$, tokenize the sequence, and feed it into $f_{\text{rm}}$, which is expected to satisfy:
\begin{equation}
  \label{eq:5}
  f_{\text{rm}}(x_i,c_{i1}) >  \dots > f_{\text{rm}}(x_i,c_{iK}),
\end{equation}
where $c_{i1}$ corresponds to the ground-truth class.

Since classification lacks the rich candidate structure of generative tasks, we design a composite ranking loss with two components: $\mathcal{L}_{\text{top1}}$ and $\mathcal{L}_{\text{pairwise}}$.
The Top-1 loss enforces the ground-truth candidate to score higher than all others:
\begin{equation}
\begin{aligned}
  \label{eq:6}
  \mathcal{L}_{\text{top1}} = -\frac{1}{K-1} \sum_{j=2}^{K} \log \sigma(f_{\text{rm}}(x_i,c_{i1}) \\
  - f_{\text{rm}}(x_i,c_{ij})),
\end{aligned}
\end{equation}
where $\sigma$ is the sigmoid function.

The pairwise loss constrains the relative order among non-true candidates:
\begin{equation}
\begin{aligned}
  \label{eq:7}
  \mathcal{L}_{\text{pairwise}} = -\frac{2}{(K-1)(K-2)}  \sum_{2 \le{j} <{k} \le{K}} \\
  \log \sigma(f_{\text{rm}}(x_i,c_{ij}) - f_{\text{rm}}(x_i,c_{ik})).
\end{aligned}
\end{equation}

The overall objective is:
\begin{equation}
  \label{eq:8}
  \mathcal{L}_{\text{rm}} = \alpha \mathcal{L}_{\text{top1}} + (1-\alpha)\mathcal{L}_{\text{pairwise}},
\end{equation}
where $\alpha \in(0,1]$ balances Top-1 preference and pairwise ranking consistency.

For binary classification, the loss simplifies to:
\begin{equation}
  \label{eq:9}
  \mathcal{L}_{\text{rm}} = -\log \sigma(f_{\text{rm}}(x_i,c_{i1}) - f_{\text{rm}}(x_i,c_{i2})).
\end{equation}

This training procedure enables the RM to capture finer-grained preference signals than conventional supervision.

\subsection{Reinforcement Learning Optimization}

In the final stage, we further optimize the policy model $f_\theta$ using PPO \citep{schulman2017proximal} to improve decision quality and predictive reliability. 
A frozen reference model $f_{\text{ref}}$, with the same architecture as $f_\theta$, is used to measure deviation from the supervised baseline.

Given an input $x_i$, the policy model $f_\theta$ outputs a class distribution $p_{\theta}(y_i \, | \, x_i) $, while the reference model $f_{\text{ref}}$ outputs $p_{\text{ref}}(y_i \, | \, x_i) $.
An action $a_i$ is defined as a class label sampled from $p_{\theta}(y_i \, | \, x_i) $, and its raw reward $f_{\text{rm}}(x_i,a_i)$ is computed by the RM.
To prevent the policy from drifting excessively from the supervised distribution, we introduce an adaptive KL-penalty \citep{ziegler2019fine} term into the reward and obtain the modified reward:
\begin{equation}
\begin{aligned}
  \label{eq:10}
  \hat{f_{\text{rm}}}(x_i,a_i) = f_{\text{rm}}(x_i,a_i) - \\
  \beta_i{D_{\text{KL}}}(p_{\theta}(\cdot \, | \, x_i)\, || \, p_{\text{ref}}(\cdot \, &| \, x_i)),
\end{aligned}
\end{equation}
where $D_{\text{KL}}$ denotes the KL divergence between the two distributions.
The KL coefficient $\beta$ is dynamically adjusted by an adaptive controller:
\begin{equation}
\begin{aligned}
  \label{eq:11}
  \beta_{i+1} = {\text{clip}}(\beta_{i} \cdot (1+\eta \\
   \,\cdot \frac{D_{\text{KL}}-D_{\text{target}}}{D_{\text{target}}}&), \beta_{\min}, \beta_{\max}),
\end{aligned}
\end{equation}
which keeps the KL divergence close to a target value $D_{\text{target}}$; $\eta$ is the update rate.

During PPO optimization, ClaHF maximizes the following clipped surrogate objective for each sample by comparing the probability ratio between the current and old policies:
\begin{equation}
\begin{aligned}
  \label{eq:12}
  \mathcal{L}_{\text{policy}} = \mathbb{E}_{x_i \sim \mathcal{D}} [-\min(r_iA_i, \\ 
  \text{clip}(r_i,1-\epsilon,1+\epsilon)&A_i)],
\end{aligned}
\end{equation}
where $r_i = \frac{p_{\theta}(a_i \, | \, x_i)}{p_{\text{ref}}(a_i \, | \, x_i)}$ is the probability ratio of the new and old policies, and $A_i$ is the advantage function defined as:
\begin{equation}
  \label{eq:13}
  A_i = \hat{f_{\text{rm}}}(x_i,a_i) - V(x_i),
\end{equation}
with $V$ denoting the output of the value function network.
The value function is introduced to reduce training variance and stabilize policy learning.
It is optimized using mean squared error:
\begin{equation}
\begin{aligned}
  \label{eq:14}
  \mathcal{L}_{\text{value}} = \mathbb{E}_{x_i \sim \mathcal{D}} [(V(x_i)-\hat{f_{\text{rm}}}(x_i,a_i))^2].
\end{aligned}
\end{equation}

The final optimization objective in the RL stage is:
\begin{equation}
  \label{eq:15}
  \mathcal{L}_{\text{rl}} = \mathcal{L}_{\text{policy}} + \gamma \mathcal{L}_{\text{value}},
\end{equation}
where $\gamma$ controls the relative weight of the value loss.

\begin{table*}[t]
\setlength\tabcolsep{3pt}
\centering
\small
\begin{tabular}{cc|ccc|ccc|ccc}
\toprule
\multirow{2}{*}{Backbone} & \multirow{2}{*}{Method} & \multicolumn{3}{c}{AG News} & \multicolumn{3}{c}{SST-5} & \multicolumn{3}{c}{Emotion} \\
\cmidrule(lr){3-5} \cmidrule(lr){6-8} \cmidrule(lr){9-11}
& & Acc ($\uparrow$) & F1 ($\uparrow$) & ECE ($\downarrow$) & Acc ($\uparrow$) & F1 ($\uparrow$) & ECE ($\downarrow$) & Acc ($\uparrow$) & F1 ($\uparrow$) & ECE ($\downarrow$) \\
\midrule
\multirow{2}{*}{BERT} & SFT & 91.91$_{\pm 0.0}$ & 91.90$_{\pm 0.0}$ & 8.35$_{\pm 0.3}$ & 52.49$_{\pm 0.1}$ & 51.28$_{\pm 0.1}$ & 13.44$_{\pm 0.7}$ & 92.25$_{\pm 0.2}$ & 87.71$_{\pm 0.5}$ & 6.51$_{\pm 0.4}$ \\
& ClaHF & 92.18$_{\pm 0.1}$ & 92.17$_{\pm 0.1}$ & 6.96$_{\pm 0.5}$ & 54.30$_{\pm 0.1}$ & 51.50$_{\pm 0.8}$ & 11.36$_{\pm 0.3}$ & 92.85$_{\pm 0.1}$ & 88.05$_{\pm 0.4}$ & 5.74$_{\pm 0.5}$ \\
\multirow{2}{*}{RoBERTa} & SFT & 94.44$_{\pm 0.0}$ & 94.44$_{\pm 0.0}$ & 7.47$_{\pm 0.3}$ & 54.35$_{\pm 0.7}$ & 52.71$_{\pm 0.9}$ & 10.39$_{\pm 0.5}$ & 92.55$_{\pm 0.1}$ & 88.40$_{\pm 0.4}$ & 7.23$_{\pm 0.1}$ \\
& ClaHF & 94.61$_{\pm 0.0}$ & 94.60$_{\pm 0.0}$ & 5.52$_{\pm 0.2}$ & 55.87$_{\pm 0.4}$ & 54.17$_{\pm 0.6}$ & 8.43$_{\pm 0.3}$ & 92.93$_{\pm 0.1}$ & 88.93$_{\pm 0.1}$ & 6.80$_{\pm 0.1}$ \\
\multirow{2}{*}{T5} & SFT & 94.51$_{\pm 0.2}$ & 94.51$_{\pm 0.1}$ & 5.27$_{\pm 0.2}$ & 56.19$_{\pm 0.4}$ & 53.11$_{\pm 0.3}$ & 12.09$_{\pm 0.5}$ & 91.80$_{\pm 0.1}$ & 87.11$_{\pm 0.4}$ & 6.19$_{\pm 0.3}$ \\
& ClaHF & 94.79$_{\pm 0.1}$ & 94.79$_{\pm 0.1}$ & 4.52$_{\pm 0.1}$ & 57.47$_{\pm 0.8}$ & 54.72$_{\pm 0.9}$ & 11.02$_{\pm 0.4}$ & 92.20$_{\pm 0.1}$ & 88.11$_{\pm 0.1}$ & 4.74$_{\pm 0.3}$ \\
\multirow{2}{*}{OPT} & SFT & 93.77$_{\pm 0.1}$ & 93.77$_{\pm 0.1}$ & 9.60$_{\pm 0.5}$ & 52.85$_{\pm 0.4}$ & 49.72$_{\pm 0.5}$ & 10.42$_{\pm 0.8}$ & 92.28$_{\pm 0.4}$ & 87.74$_{\pm 0.7}$ & 6.62$_{\pm 0.1}$ \\
& ClaHF & 94.00$_{\pm 0.1}$ & 94.01$_{\pm 0.1}$ & 5.76$_{\pm 0.2}$ & 54.34$_{\pm 0.3}$ & 52.43$_{\pm 0.2}$ & 7.71$_{\pm 0.8}$ & 92.92$_{\pm 0.1}$ & 88.46$_{\pm 0.2}$ & 6.27$_{\pm 0.2}$ \\
\multirow{2}{*}{Qwen3} & SFT & 93.85$_{\pm 0.1}$ & 93.83$_{\pm 0.1}$ & 7.07$_{\pm 0.9}$ & 51.06$_{\pm 0.6}$ & 48.92$_{\pm 0.6}$ & 8.84$_{\pm 0.3}$ & 92.57$_{\pm 0.1}$ & 88.11$_{\pm 0.4}$ & 6.75$_{\pm 0.2}$ \\
& ClaHF & 94.17$_{\pm 0.1}$ & 94.16$_{\pm 0.1}$ & 5.22$_{\pm 0.5}$ & 52.67$_{\pm 0.1}$ & 49.55$_{\pm 1.0}$ & 7.61$_{\pm 0.4}$ & 93.62$_{\pm 0.1}$ & 90.18$_{\pm 0.1}$ & 5.06$_{\pm 0.2}$ \\
\midrule
& Average & \textbf{+0.25} & \textbf{+0.25} & \textbf{-1.96} & \textbf{+1.54} & \textbf{+1.32} & \textbf{-1.78} & \textbf{+0.61} & \textbf{+0.93} & \textbf{-0.94} \\
\bottomrule
\end{tabular}
\caption{Test performance of SFT and ClaHF on three multi-class datasets. All values are reported as the mean and standard error over five random seeds.}
\label{tab:2}
\end{table*}

Through this design, ClaHF tightly couples preference-driven reward modeling with classification decisions, continuously refining decision boundaries while maintaining stability with respect to the policy model.

\section{Experiments}

\subsection{Setups}
\label{sec:4.1}

\textbf{Datasets.} We select eight datasets to evaluate the effectiveness of ClaHF from three perspectives: binary classification, multi-class classification, and software engineering code tasks.
Specifically, the binary classification tasks include: (1) SST-2 \citep{socher2013recursive}, (2) CoLA \citep{warstadt2019neural}, and (3) MRPC \citep{dolan2005automatically}.
The multi-class classification tasks include: (4) AG News \citep{zhang2015character}, (5) SST-5 \citep{socher2013recursive}, and (6) Emotion \citep{saravia2018carer}.
The code tasks include: (7) Devign \citep{zhou2019devign}, and (8) BigCloneBench \citep{wang2020detecting}.
The data splits and additional details are provided in Appendix~\ref{sec:apA}.

\textbf{Baselines.} As existing RLHF frameworks are primarily designed for generative tasks and no mature solution is available for classification, we only compare ClaHF with standard SFT, where backbone models are trained on labeled data using cross-entropy loss.
To verify the generality of ClaHF, we adopt nine widely used pre-trained LMs as backbones, including BERT \citep{devlin2019bert}, RoBERTa \citep{liu2019roberta}, OPT \citep{zhang2022opt}, T5 \citep{raffel2020exploring}, and Qwen3 \citep{yang2025qwen3} for natural language tasks, as well as CodeBERT \citep{feng2020codebert}, CodeT5 \citep{wang2021codet5}, CodeT5+ \citep{wang2023codet5+}, and CodeGen \citep{nijkamp2022codegen} for code-related tasks.

\textbf{Implementation details.} All experiments are conducted on 8 NVIDIA RTX 4090 GPUs.
Pre-trained weights are loaded via Hugging Face \citep{wolf2020transformers}.
We use the AdamW optimizer \citep{loshchilov2017decoupled} for all experiments and train each model for 10 epochs.
During the SFT stage, the batch size is set to 16 and the learning rate to 2e-5.
For RM training, the learning rate is 1e-5 and the preference loss weight $\alpha$ is set to 0.8.
During the RL stage, the learning rate is 1e-6 and the value loss coefficient $\gamma$ is set to 0.25.
Additional hyper-parameters and training details are reported in Appendix ~\ref{sec:apB}.

\subsection{Main Results}
\label{sec:4.2}

In this section, we systematically evaluate ClaHF on multi-class, binary, and code classification tasks and compare it with the corresponding SFT baselines across a wide range of backbone models, in order to assess its cross-model and cross-task generalization ability.
To measure performance under class-imbalance scenarios, we report both Accuracy (\textit{Acc}) and macro-F1 (\textit{F1}).
For CoLA, we report the commonly used Matthews Correlation Coefficient (\textit{MCC}) \citep{chicco2021matthews}.
In addition, we introduce Expected Calibration Error (\textit{ECE}) \citep{guo2017calibration} to evaluate predictive reliability.

\subsubsection{Results on Multi-class Tasks}

\begin{table*}[t]
\setlength\tabcolsep{3pt}
\centering
\small
\begin{tabular}{cc|ccc|ccc|cc}
\toprule
\multirow{2}{*}{Backbone} & \multirow{2}{*}{Method} & \multicolumn{3}{c}{SST-2} & \multicolumn{3}{c}{MRPC} & \multicolumn{2}{c}{CoLA} \\
\cmidrule(lr){3-5} \cmidrule(lr){6-8} \cmidrule(lr){9-10}
& & Acc ($\uparrow$) & F1 ($\uparrow$) & ECE ($\downarrow$) & Acc ($\uparrow$) & F1 ($\uparrow$) & ECE ($\downarrow$) & MCC ($\uparrow$) & ECE ($\downarrow$) \\
\midrule
\multirow{2}{*}{BERT} & SFT & 89.46$_{\pm 0.24}$ & 89.43$_{\pm 0.24}$ & 9.99$_{\pm 0.14}$ & 82.18$_{\pm 0.27}$ & 79.44$_{\pm 0.96}$ & 9.84$_{\pm 0.46}$ & 55.07$_{\pm 0.34}$ & 14.30$_{\pm 0.48}$ \\
& ClaHF & 90.10$_{\pm 0.09}$ & 90.09$_{\pm 0.10}$ & 9.04$_{\pm 0.55}$ & 83.28$_{\pm 0.29}$ & 80.53$_{\pm 0.29}$ & 8.64$_{\pm 0.21}$ & 56.10$_{\pm 0.24}$ & 12.16$_{\pm 0.57}$ \\
\multirow{2}{*}{RoBERTa} & SFT & 92.57$_{\pm 0.12}$ & 92.57$_{\pm 0.12}$ & 8.24$_{\pm 0.47}$ & 87.25$_{\pm 0.18}$ & 85.57$_{\pm 0.05}$ & 6.25$_{\pm 0.29}$ & 59.73$_{\pm 0.62}$ & 11.77$_{\pm 0.29}$ \\
& ClaHF & 93.19$_{\pm 0.23}$ & 93.19$_{\pm 0.23}$ & 6.40$_{\pm 0.13}$ & 87.86$_{\pm 0.15}$ & 86.28$_{\pm 0.23}$ & 5.12$_{\pm 0.20}$ & 61.21$_{\pm 0.82}$ & 8.43$_{\pm 0.37}$ \\
\multirow{2}{*}{T5} & SFT & 93.85$_{\pm 0.23}$ & 93.85$_{\pm 0.23}$ & 7.10$_{\pm 0.60}$ & 84.58$_{\pm 0.12}$ & 81.34$_{\pm 0.27}$ & 8.15$_{\pm 0.30}$ & 58.10$_{\pm 0.48}$ & 13.58$_{\pm 0.51}$ \\
& ClaHF & 94.63$_{\pm 0.04}$ & 94.64$_{\pm 0.04}$ & 5.03$_{\pm 0.13}$ & 85.78$_{\pm 0.28}$ & 83.68$_{\pm 0.28}$ & 6.83$_{\pm 0.27}$ & 59.23$_{\pm 0.46}$ & 11.75$_{\pm 0.33}$ \\
\multirow{2}{*}{OPT} & SFT & 90.52$_{\pm 0.46}$ & 90.51$_{\pm 0.45}$ & 8.76$_{\pm 0.52}$ & 83.61$_{\pm 0.25}$ & 80.69$_{\pm 0.37}$ & 7.55$_{\pm 0.19}$ & 56.42$_{\pm 0.26}$ & 13.95$_{\pm 0.77}$ \\
& ClaHF & 90.99$_{\pm 0.23}$ & 90.99$_{\pm 0.23}$ & 6.24$_{\pm 0.31}$ & 84.69$_{\pm 0.22}$ & 81.88$_{\pm 0.85}$ & 6.44$_{\pm 0.26}$ & 57.64$_{\pm 0.35}$ & 12.90$_{\pm 0.43}$ \\
\multirow{2}{*}{Qwen3} & SFT & 90.57$_{\pm 0.16}$ & 90.57$_{\pm 0.16}$ & 8.22$_{\pm 0.36}$ & 84.41$_{\pm 0.15}$ & 81.69$_{\pm 0.54}$ & 6.85$_{\pm 0.13}$ & 56.94$_{\pm 0.17}$ & 13.68$_{\pm 0.66}$ \\
& ClaHF & 91.47$_{\pm 0.29}$ & 91.47$_{\pm 0.29}$ & 7.09$_{\pm 0.20}$ & 85.02$_{\pm 0.18}$ & 82.97$_{\pm 0.26}$ & 5.58$_{\pm 0.36}$ & 58.30$_{\pm 0.27}$ & 12.44$_{\pm 0.34}$ \\
\midrule
& Average & \textbf{+0.68}$_{\pm 0.07}$ & \textbf{+0.69}$_{\pm 0.07}$ & \textbf{-1.70}$_{\pm 0.29}$ & \textbf{+0.92}$_{\pm 0.13}$ & \textbf{+1.32}$_{\pm 0.27}$ & \textbf{-1.2}1$_{\pm 0.04}$ & \textbf{+1.24}$_{\pm 0.08}$ & \textbf{-1.92}$_{\pm 0.41}$ \\
\bottomrule
\end{tabular}
\caption{Test performance of SFT and ClaHF on three binary datasets.}
\label{tab:3}
\end{table*}

Table~\ref{tab:2} summarizes the results of ClaHF and SFT on three multi-class datasets. ClaHF consistently outperforms SFT across all backbones and datasets, showing simultaneous improvements in Acc, F1, and ECE. This indicates that incorporating preference signals effectively complements instance-wise supervision and enables the model to better distinguish decision boundaries between different classes.

On the fine-grained sentiment dataset SST-5, where semantic differences are subtle, ClaHF achieves particularly notable gains, with an average improvement of 1.54 percentage points in Acc accompanied by a significant reduction in ECE. This demonstrates that ClaHF yields stronger discriminative ability and more reliable probability calibration in complex multi-class scenarios.

Similar trends are observed on the Emotion dataset. For example, with Qwen3, ClaHF improves Acc from 92.57\% to 93.62\% while reducing ECE from 6.75 to 5.06, indicating that ClaHF not only boosts accuracy but also fundamentally promotes more reliable and well-calibrated predictive distributions.

\subsubsection{Results on Binary Tasks}

Compared with multi-class classification, binary tasks are more sensitive to decision boundaries and impose stricter requirements on model confidence and distributional consistency. Therefore, we report these results separately.

\begin{table*}[t]
\setlength\tabcolsep{5pt}
\centering
\small
\begin{tabular}{cc|ccc|ccc}
\toprule
\multirow{2}{*}{Backbone} & \multirow{2}{*}{Method} & \multicolumn{3}{c}{Devign} & \multicolumn{3}{c}{BigCloneBench}  \\
\cmidrule(lr){3-5} \cmidrule(lr){6-8}
& & Acc ($\uparrow$) & F1 ($\uparrow$) & ECE ($\downarrow$) & Acc ($\uparrow$) & F1 ($\uparrow$) & ECE ($\downarrow$)  \\
\midrule
\multirow{2}{*}{CodeBERT} & SFT & 62.15$_{\pm 0.19}$ & 58.77$_{\pm 0.50}$ & 13.48$_{\pm 0.81}$ & 97.12$_{\pm 0.14}$ & 97.11$_{\pm 0.14}$ & 4.01$_{\pm 0.43}$ \\
& ClaHF & 63.14$_{\pm 0.11}$ & 59.61$_{\pm 0.28}$ & 9.64$_{\pm 0.47}$ & 97.58$_{\pm 0.20}$ & 97.58$_{\pm 0.20}$ & 2.49$_{\pm 0.34}$ \\
\multirow{2}{*}{CodeT5} & SFT & 62.99$_{\pm 0.18}$ & 62.35$_{\pm 0.20}$ & 12.02$_{\pm 0.48}$ & 96.80$_{\pm 0.17}$ & 96.79$_{\pm 0.17}$ & 4.93$_{\pm 0.83}$ \\
& ClaHF & 63.49$_{\pm 0.36}$ & 62.50$_{\pm 0.56}$ & 9.06$_{\pm 0.24}$ & 97.29$_{\pm 0.16}$ & 97.29$_{\pm 0.16}$ & 2.46$_{\pm 0.26}$ \\
\multirow{2}{*}{CodeT5+} & SFT & 63.20$_{\pm 0.17}$ & 62.87$_{\pm 0.25}$ & 11.13$_{\pm 0.42}$ & 97.87$_{\pm 0.10}$ & 97.87$_{\pm 0.10}$ & 2.08$_{\pm 0.06}$ \\
& ClaHF & 64.04$_{\pm 0.15}$ & 63.57$_{\pm 0.14}$ & 8.60$_{\pm 0.53}$ & 98.14$_{\pm 0.07}$ & 98.13$_{\pm 0.07}$ & 1.79$_{\pm 0.10}$ \\
\multirow{2}{*}{CodeGen} & SFT & 62.38$_{\pm 0.29}$ & 59.29$_{\pm 0.24}$ & 15.11$_{\pm 0.44}$ & 97.60$_{\pm 0.14}$ & 97.60$_{\pm 0.14}$ & 2.43$_{\pm 0.27}$ \\
& ClaHF & 63.45$_{\pm 0.11}$ & 61.95$_{\pm 0.87}$ & 11.13$_{\pm 0.71}$ & 97.98$_{\pm 0.09}$ & 97.98$_{\pm 0.09}$ & 1.92$_{\pm 0.05}$ \\
\midrule
& Average & \textbf{+0.85}$_{\pm 0.12}$ & \textbf{+1.09}$_{\pm 0.55}$ & \textbf{-3.33}$_{\pm 0.35}$ & \textbf{+0.40}$_{\pm 0.05}$ & \textbf{+0.40}$_{\pm 0.05}$ & \textbf{-1.20}$_{\pm 0.50}$ \\
\bottomrule
\end{tabular}
\caption{Test performance of SFT and ClaHF on code datasets.}
\label{tab:4}
\end{table*}

As shown in Table~\ref{tab:3}, ClaHF consistently outperforms SFT on SST-2, MRPC, and CoLA across all backbones, with particularly large gains in ECE. On SST-2, ClaHF achieves simultaneous improvements in Acc and F1, demonstrating that preference signals help the model better capture the relative ordering between positive and negative samples.

The advantage of ClaHF is even more pronounced on MRPC, which requires fine-grained judgments of semantic equivalence and is highly sensitive to boundary cases. Compared with SFT, ClaHF substantially improves performance and reduces ECE by about 1.21, indicating enhanced decision stability near hard examples. On CoLA, ClaHF improves MCC by approximately 1.24 while reducing ECE by about 1.92, further confirming its dual benefits in refining decision boundaries and improving confidence calibration.

\subsubsection{Results on Code Tasks}

Compared with natural language, source code is a structured and formal language with strict syntactic and semantic constraints, which places higher demands on model discrimination and stability.

Table~\ref{tab:4} presents the results on Devign and BigCloneBench. Overall, ClaHF achieves consistent improvements across all code backbones. On Devign, ClaHF reduces ECE by an average of 3.33 while also yielding small but consistent gains in Acc and F1, demonstrating its effectiveness on structured code data. On BigCloneBench, ClaHF again shows stable advantages, especially in terms of ECE, indicating that it maintains reliable confidence calibration even in high-accuracy regimes—an essential property for practical software engineering applications.

\subsection{Ablation Study}
\label{sec:4.3}

To further investigate the contribution of each stage in ClaHF, we conduct ablation studies from two perspectives. Unless otherwise specified, all experiments in this section are conducted on the Emotion dataset \citep{saravia2018carer} with the Qwen3  \citep{yang2025qwen3} backbone, and all other settings are kept identical.

\begin{table*}[t]
\setlength\tabcolsep{6pt}
\centering
\small
\begin{tabular}{c|ccc|ccc}
\toprule
Method & $\mathcal{L}_{\text{sft}}$ & $\mathcal{L}_{\text{top1}}$ & $\mathcal{L}_{\text{pairwise}}$ & Acc ($\uparrow$) & F1 ($\uparrow$) & ECE ($\downarrow$) \\
\midrule
SFT & $\checkmark$ & \textendash & \textendash & 92.57$_{\pm 0.14}$ & 88.11$_{\pm 0.40}$ & 6.75$_{\pm 0.22}$ \\
ClaHF-Pair & $\checkmark$ & \textendash & $\checkmark$ & 92.78$_{\pm 0.14}$ & 88.88$_{\pm 0.12}$ & 6.44$_{\pm 0.21}$ \\
ClaHF-Top1 & $\checkmark$ & $\checkmark$ & \textendash &  93.15$_{\pm 0.20}$ & 89.28$_{\pm 0.40}$ & 5.98$_{\pm 0.15}$ \\
ClaHF (Full) & $\checkmark$ & $\checkmark$ & $\checkmark$ & \textbf{93.62}$_{\pm 0.10}$ & \textbf{90.18}$_{\pm 0.12}$ & \textbf{5.06}$_{\pm 0.21}$ \\
\bottomrule
\end{tabular}
\caption{Effect of RM training objectives on classification performance.}
\label{tab:5}
\end{table*}

\begin{figure*}[t]
  \centering
  \begin{subfigure}{0.3\linewidth}
    \centering
    \includegraphics[width=\linewidth]{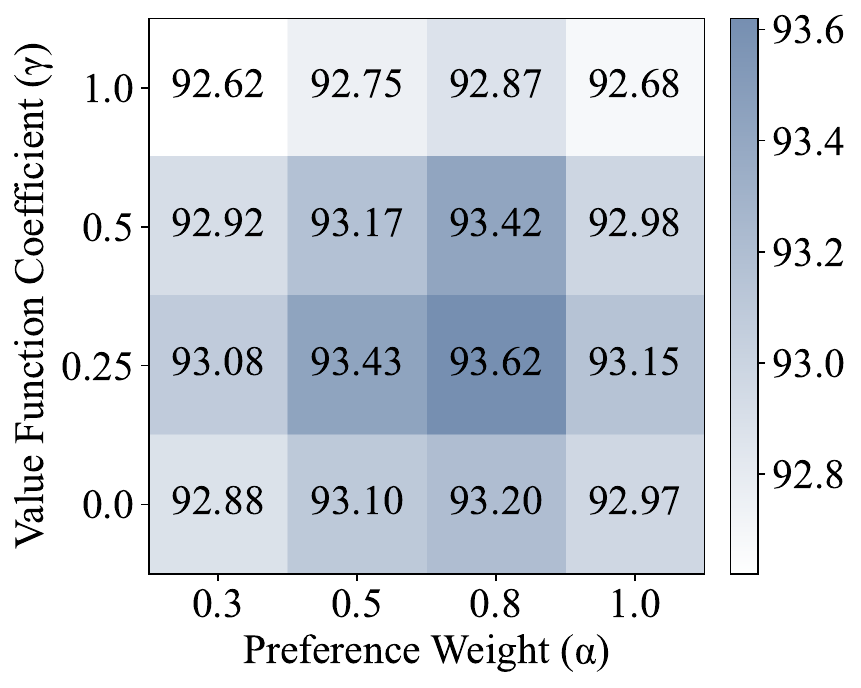}
    \caption{Heatmap of Acc.}
    \label{fig:3-a}
  \end{subfigure}
  \hfill
  \begin{subfigure}{0.3\linewidth}
    \centering
    \includegraphics[width=\linewidth]{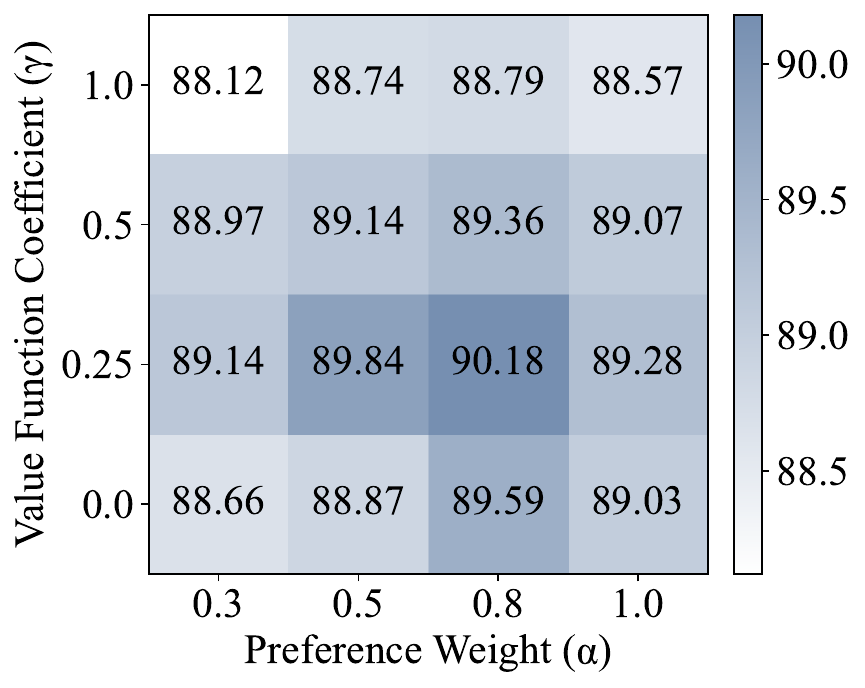}
    \caption{Heatmap of F1.}
    \label{fig:3-b}
  \end{subfigure}
  \hfill
  \begin{subfigure}{0.3\linewidth}
    \centering
    \includegraphics[width=\linewidth]{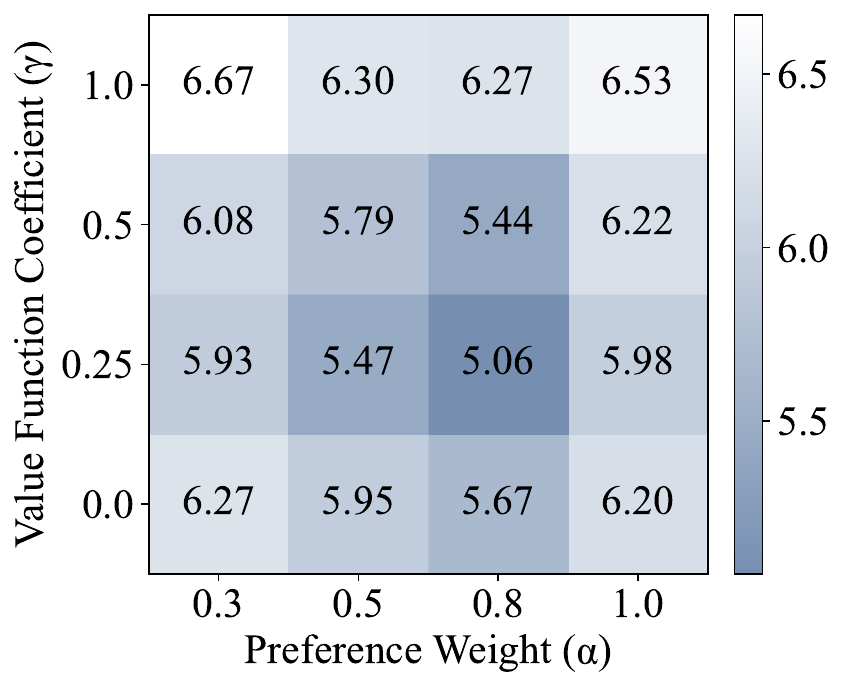}
    \caption{Heatmap of ECE.}
    \label{fig:3-c}
  \end{subfigure}
  \caption{Performance heatmaps of ClaHF in terms of Acc, F1, and ECE over a continuous $\alpha \times \gamma$ grid.}
  \label{fig:3}
\end{figure*}

\subsubsection{Effect of RM Training Objectives}

Table~\ref{tab:5} reports the impact of different RM training objectives on the final classification performance. In this group of experiments, we only vary the loss components used in the RM stage. To isolate the effect of preference signals, we construct two variants of ClaHF: (1) ClaHF-Pair: applies equal constraints to all candidates. (2) ClaHF-Top1: only enforces the margin between the Top-1 candidate and the remaining candidates.

As shown in the table, all three ClaHF variants outperform SFT, confirming that preference signals effectively complement single-label supervision. However, ClaHF-Pair yields only limited improvements, indicating that enforcing pairwise ranking consistency alone is insufficient to provide strong discriminative signals and behaves similarly to conventional learning-to-rank methods. 

When both Top-1 and pairwise constraints are combined, the model achieves the best performance across Acc, F1, and ECE, which significantly exceeds SFT and the two variants. This result indicates that the two types of preference signals are complementary: the Top-1 constraint emphasizes discriminability of the optimal decision, while the pairwise constraint preserves the global ranking structure. Additional studies are reported in Appendix~\ref{sec:apC}.

\subsubsection{Stability Analysis}

We further analyze the joint effect of the preference weight $\alpha$ and the value coefficient $\gamma$ over a continuous hyperparameter space to assess the stability of ClaHF and the structure of its optimal region.

Figure~\ref{fig:3} presents heatmaps of Acc, F1, and ECE over a continuous $\alpha \times \gamma$ grid. The performance surfaces vary smoothly rather than exhibiting abrupt oscillations, indicating that ClaHF is robust to hyperparameter perturbations and that small changes within reasonable ranges do not lead to dramatic performance drops.

Moreover, the optimal performance does not concentrate at a single point but spans a coherent region. Specifically, when $\alpha \in [0.7,0.9]$ and $\gamma \in [0.2,0.3]$, the model achieves near-optimal Acc and F1 while maintaining low ECE. The existence of this structured optimal region demonstrates that the improvements brought by ClaHF are not the result of accidental tuning, but remain stable across a range of reasonable configurations.

\section{Conclusion}

We propose ClaHF, a human-feedback-inspired RL framework for text classification, which for the first time systematically introduces the complete RLHF pipeline into classification tasks. ClaHF is tailored to the decision structure of classification and incorporates dedicated designs for preference data construction, reward modeling, and policy optimization, enabling RL to effectively optimize discriminative prediction objectives. Compared with SFT, ClaHF achieves stable and consistent improvements in both performance and calibration across a wide range of natural language and code classification tasks, demonstrating that preference-driven RL is also a powerful paradigm for enhancing the effectiveness and reliability of classification models.

\section*{Limitations}

Despite the promising results of ClaHF on multiple text classification benchmarks, it has two main limitations. (1) ClaHF relies on automatically constructed preference data derived from original labels. Although this avoids additional human annotation costs, the quality of RM training is inevitably constrained by the noise and bias in the original datasets. Integrating ClaHF with real human feedback is a promising direction to further improve performance and robustness. (2) Our experiments focus on single-label text classification. The applicability of ClaHF to more complex discriminative settings, such as multi-label and hierarchical classification, remains unexplored and is left for future work.

\section*{Acknowledgments}

% Bibliography entries for the entire Anthology, followed by custom entries
%\bibliography{custom,anthology-overleaf-1,anthology-overleaf-2}

% Custom bibliography entries only
\bibliography{custom}

\appendix

\section{Datasets}
\label{sec:apA}

As described in Section~\ref{sec:4.1}, we evaluate ClaHF on eight representative text classification benchmarks. For all datasets, we set the maximum input length to 400 tokens for tokenization. The detailed descriptions are as follows.

SST-2 \citep{socher2013recursive} is a single-sentence binary classification dataset consisting of sentences from movie reviews annotated with sentiment labels. The labels are 0 and 1, where 0 denotes negative and 1 denotes positive. The dataset contains 67,350 training samples, 873 validation samples, and 1,821 test samples. As part of the GLUE benchmark \citep{wang2018glue}, the official dataset is available at \url{https://huggingface.co/datasets/nyu-mll/glue/viewer/sst2}.

CoLA \citep{warstadt2019neural} is a single-sentence binary classification dataset for linguistic acceptability. The corpus is collected from books and journals in linguistic theory. Each sentence is annotated as to whether it is grammatically acceptable, with labels 0 and 1, where 0 indicates ungrammatical and 1 indicates grammatical. The dataset contains 8,551 training samples, 1,043 validation samples, and 1,063 test samples. It is also part of the GLUE benchmark \citep{wang2018glue} and is available at \url{https://huggingface.co/datasets/nyu-mll/glue/viewer/cola}.

MRPC \citep{dolan2005automatically} is a binary classification dataset consisting of sentence pairs extracted from online news sources, manually annotated to indicate whether the two sentences are semantically equivalent. The labels are 0 and 1, where 0 denotes not paraphrase and 1 denotes paraphrase. The dataset contains 3,668 training samples, 408 validation samples, and 1,725 test samples. It is also part of GLUE \citep{wang2018glue} and is available at \url{https://huggingface.co/datasets/nyu-mll/glue/viewer/mrpc}.

AG News \citep{zhang2015character} is a single-sentence four-class news topic classification dataset collected from global news articles. The dataset covers four major categories: World (label 0), Sports (label 1), Business (label 2), and Science/Technology (label 3). It contains 120,000 training samples and 7,600 test samples. Since no validation split is provided, we randomly sample 20\% of the training set as the validation set. The dataset is available at \url{https://huggingface.co/datasets/fancyzhx/ag_news}.

SST-5 \citep{socher2013recursive} is a fine-grained single-sentence five-class sentiment classification dataset for movie reviews. The labels range from 0 to 4, representing very negative, negative, neutral, positive, and very positive, respectively. The dataset contains 8,544 training samples, 1,101 validation samples, and 2,210 test samples. It is available at \url{https://huggingface.co/datasets/SetFit/sst5}.

Emotion \citep{saravia2018carer} is a single-sentence six-class emotion classification dataset built from English Twitter messages. It covers six emotion categories: sadness (label 0), joy (label 1), love (label 2), anger (label 3), fear (label 4), and surprise (label 5). The dataset contains 16,000 training samples, 2,000 validation samples, and 2,000 test samples. It is available at \url{https://huggingface.co/datasets/dair-ai/emotion}.

\begin{table*}
\setlength\tabcolsep{6pt}
  \centering
  \begin{tabular}{c|c|c}
    \toprule
    \textbf{Dataset $\mathcal{D}$}    & \textbf{True Label $y_i$} & \textbf{Candidate Semantic Label $c_{j}$} \\
    \midrule
    \multirow{4}{*}{AG News}
                &  0    &  Prediction: the news topic is World  \\
                &  1    &  Prediction: the news topic is Sports     \\
                &  2    &  Prediction: the news topic is Business        \\
                &  3    &  Prediction: the news topic is Science/Technology       \\
    \midrule
    \multirow{5}{*}{SST-5}
                &  0    &  Prediction: the sentence is very negative  \\
                &  1    &  Prediction: the sentence is negative     \\
                &  2    &  Prediction: the sentence is neutral        \\
                &  3    &  Prediction: the sentence is positive       \\
                &  4    &  Prediction: the sentence is very positive   \\
    \midrule
    \multirow{6}{*}{Emotion}
                &  0    &  Prediction: the emotion in the text is sadness  \\
                &  1    &  Prediction: the emotion in the text is joy     \\
                &  2    &  Prediction: the emotion in the text is love        \\
                &  3    &  Prediction: the emotion in the text is anger       \\
                &  4    &  Prediction: the emotion in the text is fear   \\  
                &  5    &  Prediction: the emotion in the text is surprise   \\ 
    \midrule
    \multirow{2}{*}{SST-2}
                &  0    &  Prediction: the sentence is negative  \\
                &  1    &  Prediction: the sentence is positive     \\    
    \midrule
    \multirow{2}{*}{CoLA}
                &  0    &  Prediction: the sentence is grammatically unacceptable  \\
                &  1    &  Prediction: the sentence is grammatically acceptable     \\   
    \midrule
    \multirow{2}{*}{MRPC}
                &  0    &  Prediction: the two sentences are not semantically equivalent  \\
                &  1    &  Prediction: the two sentences are semantically equivalent     \\  
    \midrule
    \multirow{2}{*}{Devign}
                &  0    &  Prediction: the code is non-vulnerable  \\
                &  1    &  Prediction: the code is vulnerable     \\      
    \midrule
    \multirow{2}{*}{BigCloneBench}
                &  0    &  Prediction: the two code snippets are not semantically similar  \\
                &  1    &  Prediction: the two code snippets are semantically similar     \\    
    \bottomrule
  \end{tabular}
  \caption{
    Ground-truth labels and their corresponding candidate semantic labels for the eight datasets.
  }
  \label{tab:6}
\end{table*}

Devign \citep{zhou2019devign} is a code defect detection dataset designed to identify whether a code snippet is potentially vulnerable to software attacks. The labels are 0 and 1, where 0 denotes safe code and 1 denotes vulnerable code. Following the CodeXGLUE \citep{lu2021codexglue}, we split the dataset into training, validation, and test sets with an 8:1:1 ratio, resulting in 21,854 training samples and 2,732 samples for both validation and test sets. The dataset is available at \url{https://huggingface.co/datasets/google/code_x_glue_cc_defect_detection}.

BigCloneBench \citep{wang2020detecting} is a manually annotated Java code clone detection benchmark for code clone detection and semantic similarity learning. The labels are 0 and 1, where 1 indicates that the code pair is semantically equivalent and 0 indicates non-equivalence. Since the original dataset is very large, we follow the CodeXGLUE \citep{lu2021codexglue} setting and use only 10\% of the data for training and validation. The dataset is available at \url{https://huggingface.co/datasets/google/code_x_glue_cc_clone_detection_big_clone_bench}.

\section{Additional Implementation Details}
\label{sec:apB}

In the preference data construction stage, for each ground-truth class in the eight datasets, we define a natural-language candidate semantic label, as shown in Table~\ref{tab:6}. The core motivation is to map the original discrete class label $y_i$ into a semantic description $c_{j}$ that can be directly interpreted by LMs.

In addition, during RM training, we concatenate each input sample with its candidate semantic label. Different datasets use different prefixes to explicitly distinguish task contexts and align the inputs with their semantic labels. The unified input format for the RM is $Prefix: x_i \, || \, c_{ij}$, the prefixes for different datasets are defined as follows:
\begin{itemize}
    \item AG News: News
    \item SST-5 / SST-2 / CoLA / MRPC: Sentence
    \item Emotion: Text
    \item Devign / BigCloneBench: Code
\end{itemize}

\section{Additional Ablation Studies}
\label{sec:apC}

This section further investigates ablation studies of ClaHF. Following Section~\ref{sec:4.3}, all experiments are conducted on the Emotion dataset with Qwen3 as the backbone model. All reported values are the mean and standard error over five runs with different random seeds.

\textbf{Effect of the Value Function Loss in RL.} Table~\ref{tab:7} analyzes the influence of the value loss weight in the RL stage. In these experiments, all RM settings are fixed and $\alpha$ is set to 0.8, while only the value loss coefficient $\gamma$ is varied.

\begin{table}[t]
\setlength\tabcolsep{3pt}
\centering
\small
\begin{tabular}{c|c|ccc}
\toprule
Method &  $\mathcal{L}_{\text{value}}$ & Acc ($\uparrow$) & F1 ($\uparrow$) & ECE ($\downarrow$) \\
\midrule
NoValue & \textendash & 93.20$_{\pm 0.15}$ & 89.59$_{\pm 0.45}$ & 5.67$_{\pm 0.12}$ \\
\midrule
ClaHF & $\gamma = 0.25$ & \textbf{93.62}$_{\pm 0.10}$ & \textbf{90.18}$_{\pm 0.12}$ & \textbf{5.06}$_{\pm 0.21}$ \\
ClaHF & $\gamma = 0.5$ & 93.42$_{\pm 0.13}$ & 89.36$_{\pm 0.26}$ & 5.44$_{\pm 0.12}$ \\
ClaHF & $\gamma = 1.0$ & 92.87$_{\pm 0.17}$ & 88.79$_{\pm 0.34}$ & 6.27$_{\pm 0.34}$ \\

\bottomrule
\end{tabular}
\caption{Effect of the value loss weight during RL.}
\label{tab:7}
\end{table}

Without the value loss (ClaHF-NoValue), the model still outperforms SFT, showing that preference-driven policy optimization is effective by itself. However, its performance is clearly inferior to the full ClaHF, indicating that RL without value function regularization is more susceptible to high-variance reward signals.

When a moderate value loss is introduced, the model achieves the best results on Acc, F1, and ECE, validating the crucial role of the value function in stabilizing training and smoothing policy updates. As $\gamma$ further increases, performance starts to degrade, suggesting that overly strong value constraints hinder effective preference-driven updates and thus weaken the benefits of ClaHF.

Table~\ref{tab:8} reports a comparison between ClaHF-Pair with different value function loss weights $\gamma$ and the SFT baseline. In this experiment, we keep the RM and all other PPO hyperparameters fixed and only vary the weight of the value function loss to analyze the model behavior in the absence of the Top-1 constraint, i.e., when equal constraints are imposed on all candidates without explicitly emphasizing the optimal candidate and the model is trained only to learn the relative ordering among candidates.

From the results, we observe that when the value function loss is not introduced and without the Top-1 constraint, the model can hardly outperform the SFT baseline on Acc, F1, and ECE, and even exhibits slight degradation on the calibration metric ECE. This indicates that relying solely on pairwise preference signals for policy updates makes it difficult for the model to stably learn effective discriminative boundaries.

After introducing a moderate value function loss, ClaHF-Pair achieves modest improvements over SFT, which further verifies that the value function can alleviate the instability caused by high-variance rewards in preference-based policy optimization. However, as $\gamma$ increases further, the model performance drops and even falls below that of SFT, suggesting that an excessively large value loss weight overly suppresses policy updates when the Top-1 constraint is absent.

In summary, these results demonstrate that in ClaHF-Pair, although the value function loss can improve training stability, its effective range is clearly limited and requires careful tuning. This finding also provides additional evidence for our main conclusion that Top-1 and pairwise preference signals play complementary roles in ClaHF, and that their joint modeling is crucial for stable and efficient optimization of classification policies.

\begin{table}[t]
\setlength\tabcolsep{3pt}
\centering
\small
\begin{tabular}{c|c|ccc}
\toprule
Method &  $\mathcal{L}_{\text{value}}$ & Acc ($\uparrow$) & F1 ($\uparrow$) & ECE ($\downarrow$) \\
\midrule
SFT        & \textendash & 92.57$_{\pm 0.14}$ & 88.11$_{\pm 0.40}$ & 6.75$_{\pm 0.22}$ \\
\midrule
ClaHF-Pair & \textendash & 92.53$_{\pm 0.13}$ & 88.26$_{\pm 0.23}$ & 6.92$_{\pm 0.38}$ \\
ClaHF-Pair & $\gamma = 0.25$ & 92.78$_{\pm 0.14}$ & 88.88$_{\pm 0.12}$ & 6.44$_{\pm 0.21}$ \\
ClaHF-Pair & $\gamma = 0.5$ & 92.60$_{\pm 0.10}$ & 88.57$_{\pm 0.09}$ & 6.69$_{\pm 0.28}$ \\
ClaHF-Pair & $\gamma = 1.0$ & 92.28$_{\pm 0.14}$ & 87.27$_{\pm 0.18}$ & 6.97$_{\pm 0.22}$ \\
\bottomrule
\end{tabular}
\caption{Effect of the value loss weight on ClaHF-Pair.}
\label{tab:8}
\end{table}

\textbf{Further explanation of Figure~\ref{fig:3}.} Further analysis of the interaction between $\alpha$ and $\gamma$ reveals a clear synergistic effect. When $\alpha$ is small (i.e., the RM mainly relies on relative ranking signals), varying $\gamma$ yields limited gains. In contrast, when $\alpha$ lies in the medium range, changes in $\gamma$ have a more pronounced impact on both performance and calibration, suggesting that explicit preference signals facilitate the stabilizing role of the value function. However, when $\alpha=1$, an excessively large $\gamma$ suppresses the policy’s responsiveness to preference signals and leads to performance degradation. This reveals a critical balance in ClaHF: preference guidance must be sufficiently strong to steer policy updates, while value constraints should stabilize training without over-regularization.

Finally, we observe that the distribution of ECE does not simply trade off with Acc and F1, but reaches its minimum near the same optimal region. This indicates that ClaHF simultaneously enhances discriminative performance and confidence calibration, rather than sacrificing one for the other.

\end{document}